\documentclass[11pt]{article}
\usepackage[T1]{fontenc}
\usepackage[utf8]{inputenc}
\usepackage{lmodern}
\usepackage[margin=1in]{geometry}
\usepackage{amsmath,amssymb}
\usepackage{graphicx}
\usepackage{booktabs}
\usepackage{multirow}
\usepackage{siunitx}
\usepackage[numbers,sort&compress]{natbib}
\usepackage{xcolor}
\usepackage{tikz}
\usetikzlibrary{arrows.meta,positioning,fit}
\usepackage[hidelinks]{hyperref}
\usepackage{caption}
\newcommand{\authorname}{Pedro Henrique Oliveira Silva}
\newcommand{\affiliation}{Universidade Federal de Juiz de Fora (UFJF), Juiz de Fora, MG, Brazil}
\newcommand{\authororcid}{ORCID: \href{https://orcid.org/0000-0001-5412-5675}{0000-0001-5412-5675}}

\title{Do Time-Series Foundation Models Generalize Across Load?\\
A Leakage-Safe Current-Signature Benchmark for Broken-Rotor-Bar Diagnosis}
\author{\authorname\\ \small \affiliation\\ \small \authororcid}
\date{}

\begin{document}
\maketitle

\begin{abstract}
Time-series foundation models and large language models (LLMs) are increasingly proposed for machine fault diagnosis, but most reported gains come from evaluations in which training and test windows share recordings and operating conditions. We present a leakage-safe benchmark for broken-rotor-bar diagnosis from stator current (motor current signature analysis, MCSA) on a public laboratory dataset (five severities, eight load levels, ten repetitions), in which every recording is assigned to a single split and generalization to unseen loads is tested explicitly. We compare a compact 1D CNN, the MOMENT foundation model (frozen and LoRA-adapted), an LLM fed with textual signal descriptors, and a learned bridge that maps MOMENT tokens into a small LLM (Qwen3-0.6B) answering typed multiple-choice questions. Within the training loads, nearly all methods classify every test recording correctly. Under load interpolation, the frozen foundation model is the bottleneck: the bridge on frozen MOMENT reaches 0.62 recording-level accuracy on the number of broken bars, whereas adapting MOMENT with LoRA raises the bridge to 0.82, statistically indistinguishable from the CNN (0.82; paired bootstrap difference $+0.01$, 95\% CI $[-0.04, 0.06]$), at about 500 times the inference latency and with slightly worse calibration. Extrapolating from high to low load fails for every method (0.22--0.40). A spectral analysis links this failure to physical observability: the lower sideband at $(1-2s)f$ separates healthy and four-bar rotors by \SI{17}{dB} at \SI{4}{N.m} but only \SI{2.4}{dB} at \SI{0.5}{N.m}, where it lies \SI{0.5}{Hz} from the fundamental. We also specify a complete evaluation protocol for this question, including the experiments still in progress (leave-one-load-out, a slip-aligned classical MCSA baseline, a bridge-versus-rules ablation, paraphrase consistency and per-load calibration), and mark which results are already available.
\end{abstract}

\section{Introduction}

Broken rotor bars are a classical fault of squirrel-cage induction motors. Their standard non-invasive signature is a pair of sidebands around the supply frequency $f$ in the stator current, at $(1\pm 2ks)f$, where $s$ is the slip~\citep{thomson2001mcsa,bellini2008advances}. Because $s$ grows with load, the sideband position and amplitude depend strongly on the operating point, and diagnosis at light load is known to be difficult~\citep{okwuosa2022lowload,boushaba2022picnn}.

Deep learning on such signals often reports near-perfect accuracy. Recent work shows that much of this comes from the evaluation protocol: overlapping windows from the same recording in both training and test sets, or random splits that mix operating conditions, inflate metrics substantially~\citep{vieira2025realistic,ertasgin2026leakage,yelmen2026uncertainty}, and windows with large overlap are not independent test samples~\citep{shi2026windows}. On the dataset used here, \citet{kumar2026efficientnet} report 99.8\% with vibration and mixed loads but only 76\% with current, and \citet{aitmellal2026hybrid} show that a leakage-free vibration pipeline drops to 67\% on average under leave-one-load-out.

In parallel, time-series foundation models (TSFMs) such as MOMENT~\citep{goswami2024moment}, Chronos~\citep{ansari2024chronos} and TimesFM~\citep{das2024timesfm}, and LLM-based diagnosis frameworks~\citep{tao2025llm,ren2026phmgpt,qaid2026fdllm,zou2026tfcmmfd}, are being proposed for condition monitoring. Evidence on whether TSFM representations transfer to motor faults is mixed: \citet{chen2026applicability} find that classical physical features are often more stable than Chronos or MOMENT representations on end-of-line motor tests (vibration and acoustics), while \citet{li2025dataefficient} report data-efficient motor monitoring with MOMENT and Mantis~\citep{feofanov2025mantis}.

This paper asks a narrow, testable question: \emph{under a leakage-safe protocol, do foundation-model pipelines generalize across load better than a small specialized network when diagnosing broken bars from stator current?} Our contributions are:
\begin{enumerate}
  \item A leakage-safe MCSA benchmark on a public dataset~\citep{treml2020dataset} with recording-level splits and four named scenarios (in-domain, load interpolation, load extrapolation, few labels), reporting accuracy, calibration and selective-prediction metrics at window and recording level.
  \item A comparison of a 1D CNN, frozen and LoRA-adapted MOMENT, an LLM with textual descriptors, and a MOMENT$\rightarrow$LLM bridge answering three typed questions (number of bars, severity, maintenance).
  \item Evidence that the \emph{frozen} foundation encoder is the bottleneck under load shift, and that adapting it with LoRA closes the gap to the CNN but does not surpass it.
  \item A physical analysis tying the failure of load extrapolation to the loss of sideband contrast and frequency separation at light load.
  \item A signal-to-decision interface in which a single model reads a small, fixed budget of signal tokens and answers several typed maintenance questions with calibrated option scores and an explicit abstention option.
  \item A full evaluation protocol for foundation models in MCSA under load shift (Section~\ref{sec:protocol}), with the status of each experiment stated explicitly: completed, partial, or planned.
\end{enumerate}
We also state clearly what the current evidence does \emph{not} support (Section~\ref{sec:limitations}).

\section{Related work}

\paragraph{Broken-bar diagnosis on this dataset.} The USP/UFSCar database~\citep{treml2020dataset} has been used with vibration-based transfer learning~\citep{misra2022fault} and image encodings~\citep{kumar2026efficientnet}. The closest work is \citet{aitmellal2026hybrid}, who perform a leakage-free evaluation with temporal partitioning and leave-one-load-out on the same test bench, using vibration and a CNN--LightGBM hybrid. Our study differs in modality (stator current only, the cheapest and most deployable sensor), in the models compared (foundation models and LLMs), and in reporting calibration and its relation to the MCSA physics.

\paragraph{Foundation models and LLMs for fault diagnosis.} LLM-based frameworks for bearings~\citep{tao2025llm}, prognostics~\citep{ren2026phmgpt} and explainable diagnosis~\citep{qaid2026fdllm} typically convert signals or features into text or use a learned projector into the LLM, similar in spirit to vision--language bridges such as the Q-Former~\citep{li2023blip2} and to Time-LLM~\citep{jin2024timellm}. A recent multimodal foundation model for bearings reports above 80\% accuracy on most interpolated unseen conditions~\citep{zou2026tfcmmfd}. Few of these works evaluate stator current, recording-level splits and calibration under load shift together.

\paragraph{Evaluation and calibration.} Splits by segment or condition inflate fault-diagnosis metrics~\citep{vieira2025realistic,ertasgin2026leakage}; window-level hypothesis tests with 75\% overlap have an inflated type-I error~\citep{shi2026windows}. Calibration-aware diagnosis has been studied by \citet{lin2024uncertainty} and \citet{yelmen2026uncertainty}; we use expected calibration error (ECE), temperature scaling fitted on validation data~\citep{guo2017calibration}, and the area under the risk--coverage curve (AURC)~\citep{geifman2017selective}.

\section{Data and protocol}
\label{sec:data}

\paragraph{Dataset.} The USP/UFSCar database~\citep{treml2020dataset} contains recordings of a \SI{1}{cv}, four-pole, 34-bar induction motor (rated torque \SI{4.1}{N.m}) fed from the \SI{60}{Hz} mains, with 0 to 4 broken bars (five classes), eight load torques from 0.5 to \SI{4.0}{N.m}, and ten repetitions per condition. Each recording lasts \SI{20}{s} at \SI{50}{kHz}. We use only the three stator currents; 399 of the 400 recordings are valid.

\paragraph{Pre-processing.} The phase currents are converted into an amplitude envelope around the supply frequency, in which the broken-bar modulation at $2sf$ appears as a slow oscillation. Each recording is then cut into overlapping windows of a few seconds, and each window is normalized individually, so that no statistic is estimated across recordings.

\paragraph{Leakage-safe splits.} Each recording is assigned as a whole to one split before windowing; an automated check asserts that no recording contributes windows to two splits. Four scenarios are defined:
\begin{itemize}
  \item \textbf{in-domain}: repetitions 1--7 train, 8 validation, 9--10 test, all loads (79 test recordings);
  \item \textbf{load interpolation} (\texttt{load\_interp}): train and validate on 0.5/1.5/2.5/\SI{3.5}{N.m}, test on all recordings at 1/2/3/\SI{4}{N.m} (200 test recordings); note that \SI{4}{N.m} lies slightly outside the training range;
  \item \textbf{load extrapolation} (\texttt{load\_shift}): train on loads $\ge$ \SI{2.5}{N.m}, test on 0.5 and \SI{1.0}{N.m} (99 recordings);
  \item \textbf{few labels}: in-domain with 10\% of training recordings, stratified by class.
\end{itemize}

\paragraph{Questions.} Every model answers three typed questions derived from the number of broken bars $n$: \emph{number of bars} (5 classes), \emph{severity} (none, incipient $n{=}1$, moderate $n{=}2$, severe $n\ge3$) and \emph{maintenance} (no for $n{=}0$, yes otherwise). The maintenance question is imbalanced (80\% faulty), so we also report balanced accuracy.

\paragraph{Metrics and statistics.} We report window accuracy and recording-level accuracy, where recording predictions average window log-probabilities. We also report macro-F1, mean absolute error (MAE) in number of bars, ECE (15 equal-width bins) before and after temperature scaling on the validation split, and AURC. Because overlapping windows are not independent~\citep{shi2026windows}, our comparisons use recordings as the unit: a McNemar test on discordant recordings and a bootstrap over recordings (5000 resamples) of the accuracy difference, with seed-averaged correctness per recording.

\section{Methods}
\label{sec:methods}

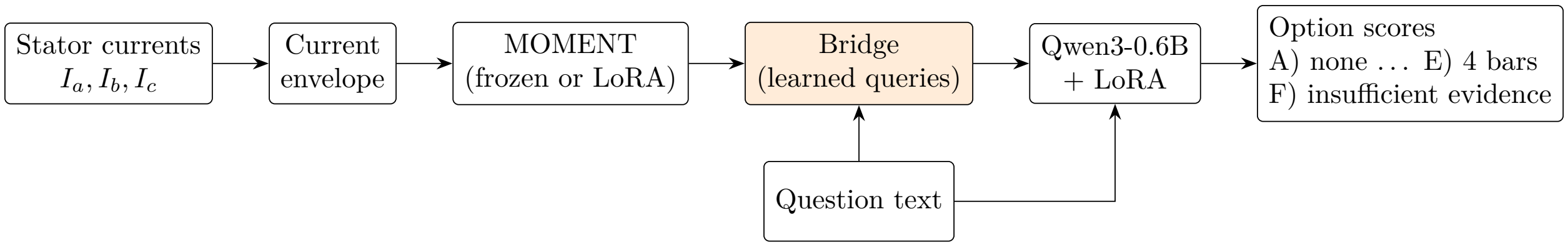
\begin{figure}[t]
\centering
\resizebox{\linewidth}{!}{%
\begin{tikzpicture}[node distance=7mm, every node/.style={font=\small}, box/.style={draw, rounded corners=2pt, align=center, minimum height=10mm, inner sep=4pt}, ar/.style={-{Stealth[length=2mm]}}]
\node[box] (sig) {Stator currents\\$I_a, I_b, I_c$};
\node[box, right=of sig] (env) {Current\\envelope};
\node[box, right=of env] (enc) {MOMENT\\(frozen or LoRA)};
\node[box, right=of enc, fill=orange!15] (br) {Bridge\\(learned queries)};
\node[box, right=of br] (llm) {Qwen3-0.6B\\+ LoRA};
\node[box, right=of llm, align=left] (out) {Option scores\\A) none \dots\ E) 4 bars\\F) insufficient evidence};
\node[box, below=7mm of br] (q) {Question text};
\draw[ar] (sig) -- (env); \draw[ar] (env) -- (enc); \draw[ar] (enc) -- (br); \draw[ar] (br) -- (llm); \draw[ar] (llm) -- (out);
\draw[ar] (q) -- (br); \draw[ar] (q) -| (llm);
\end{tikzpicture}}
\caption{Bridge pipeline. Encoder tokens are summarized by learned queries conditioned on the question and injected as a fixed number of soft tokens into a small LLM, which scores the answer options, including an abstention option. The CNN and linear baselines use the same envelope windows.}
\label{fig:pipeline}
\end{figure}

\paragraph{B1: 1D CNN.} A compact convolutional network on the envelope windows (about 0.2M parameters), one model per question, in the style of networks commonly used for raw-signal fault diagnosis~\citep{zhang2017wdcnn}.

\paragraph{B2: frozen MOMENT + linear.} The MOMENT foundation model~\citep{goswami2024moment} embeds each window with frozen weights; a linear classifier per question is trained on the embeddings.

\paragraph{MOMENT-LoRA + linear.} The same encoder adapted with low-rank adaptation (LoRA)~\citep{hu2022lora} jointly with linear heads for the three questions, using only the training split of each scenario.

\paragraph{B4: LLM with textual descriptors.} The same small LLM (Qwen3-0.6B~\citep{qwen3}) adapted with LoRA receives a prompt in which a few spectral and statistical descriptors of the window replace the signal tokens.

\paragraph{Bridge.} In the spirit of the Q-Former~\citep{li2023blip2}, a small set of learned queries reads the encoder tokens together with the question and produces a fixed number of soft tokens, which are inserted into the LLM input. The LLM, adapted with LoRA, scores the answer options; training uses a cross-entropy loss over the options and a regularizer that keeps the adapted LLM close to the base model. Only the bridge and the LLM adapter are trained (about 7M parameters); the encoder is either the frozen MOMENT or the MOMENT-LoRA encoder above, frozen after adaptation.

\paragraph{Typed decision interface.} The bridge is designed as an interface between a signal encoder and a decision, rather than as a free-text generator. (i) \emph{Fixed signal budget:} whatever the number of channels or patches, the LLM reads a small fixed number of soft tokens, so the cost of the language model does not grow with the signal length. (ii) \emph{Question-conditioned reading:} the queries attend jointly to the signal tokens and to the question, so the same window can be summarized differently for different questions. (iii) \emph{Typed output:} each question has a closed option set, and the answer is the normalized distribution of the LLM over the option letters, not generated text; this gives probabilities that can be calibrated and evaluated with ECE and risk--coverage curves. (iv) \emph{One model, several questions:} a single bridge and LoRA serve the three questions, and new questions with their own option sets can be added without changing the architecture. (v) \emph{Abstention:} the ``insufficient evidence'' option and the confidence on the chosen option support selective prediction.

All models were trained on a single NVIDIA RTX 2060 (6\,GB) in mixed precision. Unless noted, results are averaged over 5 seeds (CNN, B2, B4, frozen-encoder bridge) or 3 seeds (MOMENT-LoRA variants).

\section{Results}
\label{sec:results}

\subsection{Within the training loads, the task is easy}
In the in-domain scenario every method except B4 classifies all 79 test recordings correctly for all three questions (Table~\ref{tab:scenarios}). Window accuracy on the number of bars ranges from 0.85 (bridge) to 0.96 (CNN). Distinguishing methods therefore requires a distribution shift.

\begin{table}[t]
\centering
\caption{Number of broken bars across scenarios: window accuracy / recording accuracy, and window ECE. Mean over seeds (5 unless marked $^\dagger$, 3 seeds). Chance is 0.20.}
\label{tab:scenarios}
\small
\resizebox{\linewidth}{!}{\begin{tabular}{lcccc}
\toprule
Method & in-domain & load interp. & load extrap. & few labels (10\%) \\
\midrule
B1 CNN                      & 0.96 / 1.00 \;(0.04) & 0.77 / 0.82 \;(0.13) & 0.22 / 0.22 \;(0.71) & 0.68 / 0.78 \;(0.10) \\
B2 MOMENT frozen + linear   & 0.91 / 1.00 \;(0.07) & 0.56 / 0.59 \;(0.26) & 0.28 / 0.29 \;(0.61) & 0.52 / 0.61 \;(0.10) \\
MOMENT-LoRA + linear$^\dagger$ & 0.91 / 1.00 \;(0.03) & 0.74 / 0.79 \;(0.13) & -- & -- \\
B4 LLM + descriptors        & 0.86 / 0.98 \;(0.06) & 0.50 / 0.60 \;(0.29) & 0.35 / 0.40 \;(0.47) & 0.48 / 0.58 \;(0.10) \\
Bridge, frozen MOMENT       & 0.85 / 1.00 \;(0.01) & 0.55 / 0.62 \;(0.29) & 0.29 / 0.30 \;(0.26) & 0.45 / 0.52 \;(0.11) \\
Bridge, MOMENT-LoRA$^\dagger$ & -- & 0.78 / 0.82 \;(0.15) & -- & -- \\
\bottomrule
\end{tabular}}
\end{table}

\subsection{Under load interpolation, the frozen encoder is the bottleneck}
Table~\ref{tab:interp} details the load-interpolation scenario. With the frozen encoder, the bridge (0.62 recording accuracy) is no better than a linear probe on the same embeddings (0.59; difference $+0.035$, 95\% CI $[-0.03, 0.10]$): the LLM adds little on top of a representation that does not separate the classes at unseen loads. Adapting MOMENT with LoRA lifts the linear probe from 0.59 to 0.79 and the bridge from 0.62 to 0.82 (paired difference over the same 200 recordings and seeds 0--2: $+0.19$, CI $[+0.14, +0.24]$, McNemar $p<10^{-20}$). The adapted bridge is statistically indistinguishable from the CNN: $+0.01$, CI $[-0.04, +0.06]$, McNemar $p=0.66$; it also has the lowest MAE in number of bars (0.20 vs.\ 0.34 for the CNN), i.e.\ its errors are more often off by one bar. The CNN remains better calibrated (ECE 0.13 vs.\ 0.15) and is better on balanced accuracy for the maintenance question (0.92 vs.\ 0.83); the bridge's higher plain accuracy on maintenance (0.92 vs.\ 0.86) reflects the 80\% prior of faulty recordings and should not be read as an advantage.

\begin{table}[t]
\centering
\caption{Load interpolation (train 0.5/1.5/2.5/3.5\,N$\cdot$m, test 1/2/3/4\,N$\cdot$m, 200 test recordings). Number of bars unless noted. Rec.: recording level. Maint.\ b-acc: balanced accuracy on the maintenance question.}
\label{tab:interp}
\small
\resizebox{\linewidth}{!}{\begin{tabular}{lccccccc}
\toprule
Method & seeds & Window acc & Rec.\ acc & Rec.\ MAE & ECE & Severity acc & Maint.\ b-acc \\
\midrule
B1 CNN                    & 5 & 0.77 $\pm$ 0.08 & 0.82 $\pm$ 0.06 & 0.34 & \textbf{0.13} & \textbf{0.82} & \textbf{0.92} \\
B2 MOMENT frozen + linear & 5 & 0.56 $\pm$ 0.00 & 0.59 $\pm$ 0.01 & 0.53 & 0.26 & 0.60 & 0.84 \\
MOMENT-LoRA + linear      & 3 & 0.74 $\pm$ 0.04 & 0.79 $\pm$ 0.04 & 0.22 & \textbf{0.13} & 0.79 & 0.78 \\
B4 LLM + descriptors      & 5 & 0.50 $\pm$ 0.04 & 0.60 $\pm$ 0.07 & 0.51 & 0.29 & 0.61 & 0.79 \\
Bridge, frozen MOMENT     & 5 & 0.55 $\pm$ 0.02 & 0.62 $\pm$ 0.03 & 0.45 & 0.29 & 0.63 & 0.76 \\
Bridge, MOMENT-LoRA       & 3 & \textbf{0.78 $\pm$ 0.03} & \textbf{0.82 $\pm$ 0.06} & \textbf{0.20} & 0.15 & 0.81 & 0.83 \\
\bottomrule
\end{tabular}}
\end{table}

\begin{figure}[t]
\centering
\includegraphics[width=\linewidth]{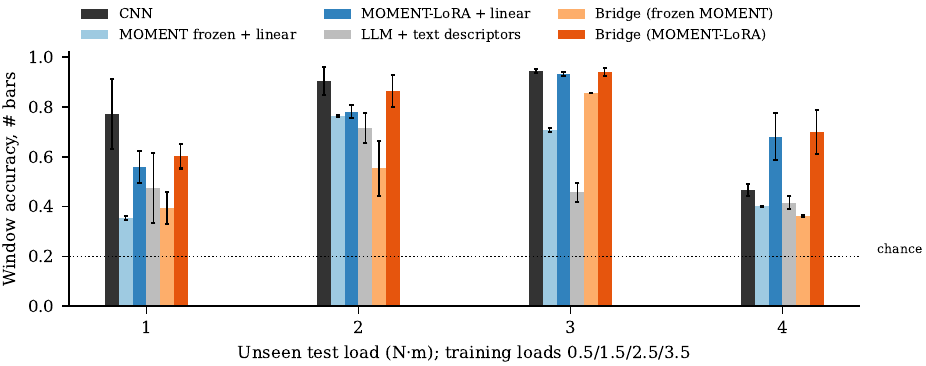}
\caption{Window accuracy on the number of broken bars per unseen test load in the load-interpolation scenario (seeds 0--2, error bars: s.d.\ over seeds).}
\label{fig:perload}
\end{figure}

Figure~\ref{fig:perload} breaks accuracy down by test load. At 2 and \SI{3}{N.m} the CNN and both LoRA-adapted models exceed 0.85. At \SI{1}{N.m} all methods degrade, the CNN least. At \SI{4}{N.m}, just above the highest training load, the CNN falls to 0.47 while both MOMENT-LoRA variants stay near 0.70: the adapted foundation encoder extrapolates upward in load better than the CNN, even though both are similar on average.

\subsection{Extrapolating to light load fails for every method}
When trained on loads $\ge$ \SI{2.5}{N.m} and tested at 0.5 and \SI{1.0}{N.m}, all methods are close to chance on the number of bars (recording accuracy 0.22--0.40, Table~\ref{tab:scenarios}). The CNN is the most confidently wrong (ECE 0.71); the frozen-encoder bridge has the lowest ECE (0.26), but this reflects lower confidence when every method fails rather than better calibration in general: in the load-interpolation scenario, where the CNN is accurate, the CNN is also the best calibrated. Neither unsupervised adaptation of batch-normalization statistics (AdaBN~\citep{li2018adabn}) nor feeding the estimated slip as an extra input improved the number-of-bars accuracy in this scenario in preliminary three-seed runs (CNN 0.23 baseline, 0.23 AdaBN, 0.21 with slip).

\subsection{Physical observability explains the asymmetry}
\label{sec:physics}
\begin{figure}[t]
\centering
\includegraphics[width=\linewidth]{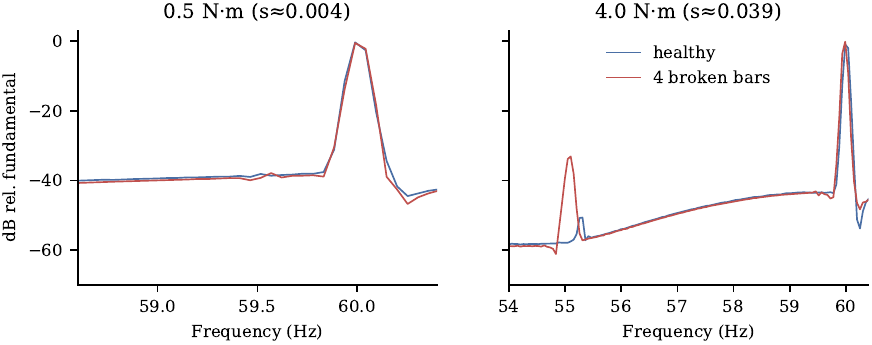}
\caption{Phase-A current spectrum near the fundamental (Hann window, full recording, averaged over the ten repetitions). At \SI{4}{N.m} the four-bar rotor shows a clear lower sideband near \SI{55}{Hz}; at \SI{0.5}{N.m} the sideband sits about \SI{0.5}{Hz} below \SI{60}{Hz} and is barely distinguishable from the healthy rotor. Note the different frequency spans.}
\label{fig:physics}
\end{figure}

\begin{figure}[t]
\centering
\includegraphics[width=\linewidth]{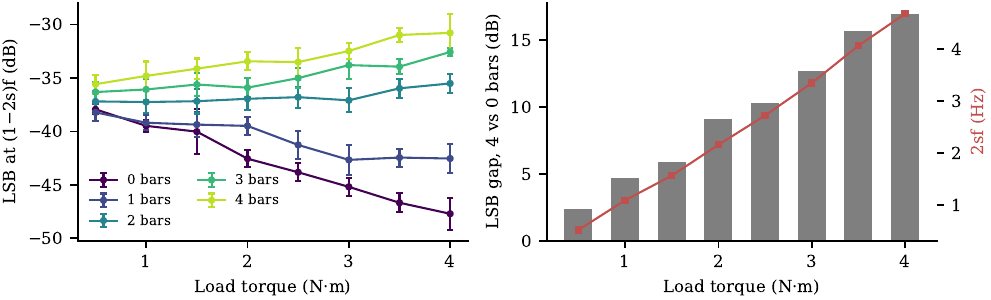}
\caption{Left: lower-sideband amplitude at $(1-2s)f$ relative to the fundamental, per number of broken bars and load (mean $\pm$ s.d.\ over repetitions). Right: gap between four-bar and healthy rotors (bars) and modulation frequency $2sf$ (line) versus load.}
\label{fig:sideband}
\end{figure}

We estimated the slip of each recording from the current itself, using the full record, and measured the lower sideband amplitude at $(1-2s)f$ relative to the fundamental. The slip grows from 0.004 at \SI{0.5}{N.m} to 0.039 at \SI{4}{N.m}, so the sideband moves from \SI{0.5}{Hz} to \SI{4.7}{Hz} below the fundamental (Fig.~\ref{fig:physics}). The contrast between healthy and four-bar rotors grows monotonically with load, from \SI{2.4}{dB} at \SI{0.5}{N.m} to \SI{16.9}{dB} at \SI{4}{N.m} (Fig.~\ref{fig:sideband}). At light load the signature is weak \emph{and} so close to the fundamental that a window of a few seconds contains only a few periods of the $2sf$ modulation. A model trained at high load has learned a modulation that is both stronger and faster than anything present at light load; extrapolation then fails regardless of the architecture. Interpolation succeeds because the test loads are bracketed by training loads with similar contrast and modulation frequency. We present this as an association consistent with MCSA physics, not as a causal decomposition; separating the effects of resolution, signal-to-noise ratio and inter-repetition variability is future work.

\subsection{Bridge size}
In a preliminary ablation in the in-domain scenario, changing the number of bridge queries over an eightfold range had little effect: window accuracy on the number of bars stayed between 0.85 and 0.86 and window ECE between 0.008 and 0.012 (1 to 5 seeds per setting). Since every variant classifies all in-domain test recordings correctly, this ablation must be repeated under load shift before drawing conclusions; the remaining planned ablations of the bridge have not been run.

\subsection{Cost}
The CNN trains in about \SI{18}{s} per seed and infers in \SI{0.02}{ms} per window on the RTX 2060. The bridge requires a MOMENT-LoRA adaptation (about \SI{20}{min}, 2.1\,GB peak memory) and bridge training (about \SI{28}{min}, 3.1\,GB), and infers in \SI{13}{ms} per window, about 500 times slower. Accuracy parity at load interpolation therefore comes at a cost several orders of magnitude higher.

\section{Evaluation protocol and experiments in progress}
\label{sec:protocol}
The results above are a first instance of a broader protocol for evaluating foundation models and LLM interfaces in MCSA under load shift. We describe it in full here so that the methodology is fixed before the remaining results are known. Table~\ref{tab:protocol} lists each component and its status in this version; components marked \emph{planned} have no results in this version, and \emph{partial} ones are reported only for the scenarios and seeds stated.

\begin{table}[t]
\centering
\caption{Evaluation protocol. Status in this version: \emph{done} (results reported above), \emph{partial} (some scenarios or seeds), \emph{planned} (specified, no results yet).}
\label{tab:protocol}
\small
\begin{tabular}{p{0.30\linewidth}p{0.42\linewidth}p{0.18\linewidth}}
\toprule
Component & Purpose & Status \\
\midrule
Recording-level splits, automated leakage check & Prevent overlapping windows from the same recording in train and test & done \\
Scenarios: in-domain, load interpolation, load extrapolation, few labels & Separate in-distribution accuracy from generalization across load & done \\
Recording-level McNemar and bootstrap & Valid inference with overlapping windows & done (load interp.) \\
Calibration: ECE, temperature scaling on validation, AURC & Measure confidence quality and selective prediction & done \\
MOMENT-LoRA encoder and bridge & Test whether the frozen encoder is the bottleneck & partial (load interp., 3 seeds) \\
Bridge on MOMENT-LoRA in all scenarios, 5 seeds & Complete Tables~\ref{tab:scenarios}--\ref{tab:interp} & running \\
Ordinal and per-class metrics (balanced accuracy, MAE, confusion) & Distinguish off-by-one errors from gross errors & partial \\
Accuracy and confidence per load & Link performance and calibration to sideband contrast & partial (Fig.~\ref{fig:perload}) \\
Leave-one-load-out (8 folds) & Locate where generalization breaks; compare with \citet{aitmellal2026hybrid} & planned \\
Slip-aligned classical MCSA baseline & Physical floor: sideband amplitudes at $(1\pm2s)f$ with slip estimated from the current & planned \\
Bridge vs.\ classifier + deterministic rules; frozen MOMENT + MLP & Isolate what the LLM adds over a classifier with a fixed mapping to severity and maintenance & planned \\
Consistency across paraphrased questions & Check that answers to equivalent questions agree with each other and with the predicted class & planned \\
Bridge ablations (design choices of the interface) & Identify which design choices matter under shift & partial (queries, in-domain) \\
Fair cost table & Parameters, peak memory, training and inference time on the same GPU & partial \\
Controlled spectral analysis (window length, SNR, sideband position) & Separate resolution and SNR effects at light load & planned \\
Ten seeds; second motor or dataset & Statistical power; generalization across machines & planned \\
\bottomrule
\end{tabular}
\end{table}

\paragraph{Planned components.} The classical baseline uses the slip estimated from the current to read the sideband amplitudes directly; if it matches the CNN under load interpolation, the learned models add little beyond the physics. Because severity and maintenance are deterministic functions of the number of bars in this dataset, a classifier followed by fixed rules answers them consistently by construction; the bridge is justified only if it improves on this baseline measurably, for example in calibration, abstention or robustness to paraphrased questions.

\paragraph{Extensions of the interface.} Further directions, not yet implemented, are: operating-condition context for the LLM; multimodal fusion with vibration through the same token interface; recognition of the operating regime with abstention when the signature is not observable, which our physical analysis suggests is the correct behaviour at light load; and end-to-end training of encoder, bridge and LLM. Unsupervised batch-normalization adaptation and slip as an extra CNN input, already tested, did not help under load extrapolation.

\section{Discussion}
\paragraph{What the results support.} (i) Under a recording-level protocol, the generic, frozen time-series foundation encoder does not transfer the broken-bar signature across loads as well as a small CNN trained on the same envelopes; a linear probe and an LLM bridge on top of it perform alike. (ii) Light domain adaptation of the encoder (LoRA, training split only) is what closes the gap; the LLM adds a question-answering interface and slightly smaller ordinal errors, but no significant accuracy gain over a linear head on the same adapted encoder ($+0.03$, CI $[-0.005, 0.06]$). (iii) Generalization to light load is limited by the signal itself, which bounds what any data-driven model can achieve with current-only data in this setting.

\paragraph{When might an LLM bridge be worth it?} The bridge's potential advantages lie in the interface: one model answering several typed questions, an explicit abstention option, and the possibility of combining signal evidence with text. Whether these benefits justify the cost should be measured against a classifier followed by deterministic rules (since severity and maintenance are functions of the number of bars here), and through consistency across paraphrased questions; both experiments are part of the protocol in Section~\ref{sec:protocol}.

\section{Limitations}
\label{sec:limitations}
\begin{itemize}
  \item \textbf{One test bench.} All results come from a single motor and laboratory set-up; generalization across motors is not tested.
  \item \textbf{Few seeds.} The MOMENT-LoRA variants have three seeds; the remaining methods have five. With five seeds the smallest attainable two-sided Wilcoxon $p$ across seeds is 0.0625, so we rely on recording-level tests, which pool seeds and are therefore only approximately independent.
  \item \textbf{Incomplete scenarios for the adapted encoder.} The MOMENT-LoRA bridge has so far been evaluated on load interpolation only.
  \item \textbf{Single pre-processing.} All learned models use the same envelope windows; a longer window may help light-load diagnosis. Alternative inputs that we tried (spectra and envelopes normalized by the slip) did not help.
  \item \textbf{No classical MCSA baseline yet.} A slip-aligned sideband classifier and leave-one-load-out results are needed for a direct comparison with \citet{aitmellal2026hybrid}.
  \item \textbf{Pre-training contamination.} We found no evidence that this dataset is part of MOMENT's pre-training corpus, but could not rule it out.
\end{itemize}

\section{Conclusion}
On a leakage-safe stator-current benchmark for broken rotor bars, foundation-model pipelines match but do not surpass a small CNN when the test loads are bracketed by the training loads, and only after the time-series encoder is adapted to the motor signals; a frozen encoder is the bottleneck. When the test loads are lighter than any training load, every method fails, consistent with the collapse of sideband contrast and modulation frequency at low slip. We recommend that future work on foundation models for MCSA report recording-level splits, named load-shift scenarios, calibration, and the cost relative to compact specialized baselines.

\section*{Code and data availability}
The dataset is publicly available from IEEE DataPort~\citep{treml2020dataset}. Code for pre-processing, training and evaluation is available from the author upon reasonable request.

\bibliographystyle{unsrtnat}
\bibliography{refs}

\end{document}